\documentclass[sigconf]{acmart}
\usepackage{natbib}
\usepackage{graphicx}
\usepackage{multirow}
\usepackage{amsmath}
\usepackage{booktabs}
\usepackage{subfigure}
\usepackage{epstopdf}
\usepackage{comment}
\usepackage{bbm}
\usepackage{bbold}
\usepackage{url}
\usepackage{xcolor}
\AtBeginDocument{%
  \providecommand\BibTeX{{%
    \normalfont B\kern-0.5em{\scshape i\kern-0.25em b}\kern-0.8em\TeX}}}

\copyrightyear{2021} 
\acmYear{2021} 
\setcopyright{acmlicensed}\acmConference[CIKM '21]{Proceedings of the 30th ACM International Conference on Information and Knowledge Management}{November 1--5, 2021}{Virtual Event, QLD, Australia}
\acmBooktitle{Proceedings of the 30th ACM International Conference on Information and Knowledge Management (CIKM '21), November 1--5, 2021, Virtual Event, QLD, Australia}
\acmPrice{15.00}
\acmDOI{10.1145/3459637.3482045}
\acmISBN{978-1-4503-8446-9/21/11}
\begin{document}
\fancyhead{}
\title{A Conditional Cascade  Model for Relational Triple Extraction}



\author{Feiliang Ren}
\authornote{Both authors contribute equally to this research and are listed randomly.}
\email{renfeiliang@cse.neu.edu.cn}
\orcid{0000-0001-6824-1191}
\affiliation{%
	\institution{Northeastern University}
	\city{Shenyang}
	\country{China}
}

\author{Longhui Zhang}
\affiliation{%
	\institution{Northeastern University}
	\city{Shenyang}
\country{China}
}
\authornotemark[1]

\author{Shujuan Yin}
\affiliation{%
	\institution{Northeastern University}
	\city{Shenyang}
	\country{China}
}

\author{Xiaofeng Zhao}
\affiliation{%
	\institution{Northeastern University}
	\city{Shenyang}
	\country{China}
}

\author{Shilei Liu}
\affiliation{%
	\institution{Northeastern University}
	\city{Shenyang}
	\country{China}
}

\author{Bochao Li}
\affiliation{%
	\institution{Northeastern University}
	\city{Shenyang}
	\country{China}
}

\renewcommand{\shortauthors}{F. Ren and L. Zhang, et al.}

\begin{abstract}
Tagging based methods are one of the mainstream  methods in relational triple extraction. However, most of them suffer from the class imbalance issue greatly. Here we propose a novel tagging based model that addresses this issue from following two aspects. First, at the model level, we propose a three-step extraction framework that can reduce the total number of samples greatly, which implicitly decreases the severity of the mentioned issue. Second, at the intra-model level, we propose a \emph{confidence threshold} based cross entropy loss that can directly neglect some samples in the major classes. We evaluate the proposed model on NYT and WebNLG. Extensive experiments show that it can address the mentioned issue effectively and achieves state-of-the-art results on both datasets. The source code of our model is available at:  https://github.com/neukg/ConCasRTE. 
\end{abstract}




\maketitle

\section{Introduction}
\label{section:Introduction}

Taking unstructured text (often sentences) as input,  relational triple extraction (RTE for short) aims to extract triples that are in the form of (\emph{subject, relation, object}), where both \emph{subject} and \emph{object} are entities and they are  connected semantically by \emph{relation}. RTE is  important   for some  tasks like automatic knowledge graph  construction. 

Nowadays,  the methods that jointly extract entities and relations  are dominant in RTE. Lots of novel joint extraction methods have been proposed~\cite{zheng2017joint,bekoulis2018joint,fu2019graphrel,eberts2019span,yu2019joint,nayak2020effective,yuan2020a,zeng2020copymtl,wei-etal-2020-novel}, and they achieve much better results than the  pipeline based methods. 
According to the extraction routes taken, most of existing joint extraction methods can be roughly classified into following three kinds. (i) \emph{Tagging based methods}~\cite{zheng2017joint,yu2019joint,wei-etal-2020-novel} that often use binary (positive and negative) tag   sequences   to determine: (1) the start and end tokens of entities, and (2) all  the relations for each entity pair.    (ii) 
\emph{Table-filling based methods}~\cite{wang-etal-2020-tplinker,zhang-etal-2017-end,miwa-bansal-2016-end,gupta-etal-2016-table}  that maintain a   table for each relation  and the items in  a table  usually  denotes the start and end positions of two entities (or even the types of these entities) that possess this relation. (iii)   \emph{Seq2Seq based methods}~\cite{zeng2018Extracting,zeng2020copymtl,zeng2019Learning,nayakn20}  that   view  a triple as a token sequence and  \emph{generate}  a triple in some orders,  such as first  \emph{generate} a relation, then  \emph{generate}   entities, etc.  

Recently, tagging based methods are attracting more and more research interests due to their superiority in both the  performance and the ability of extracting triples from  complex sentences that contain overlapping triples~\cite{zeng2018Extracting} or multiple triples.  
However, in these  methods, the  negative class   usually contains far more samples than the positive class   since there are always much more non-entity tokens in a sentence and most entity pairs possess only a very small number of relations. Therefore, these methods  suffer from the \emph{class imbalance issue}  greatly: the major classes (here is the negative class) have far more samples than the minor classes (here is the positive class). This issue is very harmful to performance because it makes the training inefficient and the trained model biased towards  the major classes~\cite{cui2019class,johnson2019survey}.
Most recent methods for  addressing this  issue  can be divided into following two kinds~\cite{cui2019class}: (i) re-sampling based methods\cite{zou2018unsupervised} that adjust the number of samples directly by adding repetitive data for the minor classes or removing data for the major classes; and (ii)  cost-sensitive re-weighting based methods~\cite{cui2019class,lin2020focal,menon2020long} that influence the loss function by assigning relatively higher costs to samples from minor classes. However,  as \cite{cui2019class} point out that the first ones are error-prone, and the second ones often make some assumptions on the sample difficulty and data distribution, but these assumptions do not always hold. 

Obviously, the key of addressing the  \emph{class imbalance issue} is to narrow the number gap between samples in the classes of  major   and minor.  Following this line, we propose \emph{ConCasRTE}, a  \emph{Con}ditional \emph{Cas}cade \emph{RTE} model that can address this issue existed in the tagging based RTE methods  from  following two aspects. First,    we propose  a  three-step extraction framework. Compared with 
existing two-step extraction framework~\cite{yu2019joint,wei-etal-2020-novel} that first extracts subjects then extracts objects and relations simultaneously based on the subjects extracted, this new framework   generates far less samples. Thus it narrows the mentioned number gap  implicitly due to the fact that the less  samples there are, the less possibility there would be a large mentioned  number gap. Second, we propose a  \emph{confidence threshold} based cross entropy loss function that can directly neglect lots of   samples  in the major classes, which narrows  the mentioned number gap explicitly. 
We evaluate \emph{ConCasRTE} on two benchmark datasets, namely NYT and  WebNLG. Extensive experiments show  it  is effective and  achieves the state-of-the-art results on both datasets. 

\section{Methodology}

\begin{figure}[t]
	\centering
	\includegraphics[width=0.443\textwidth]{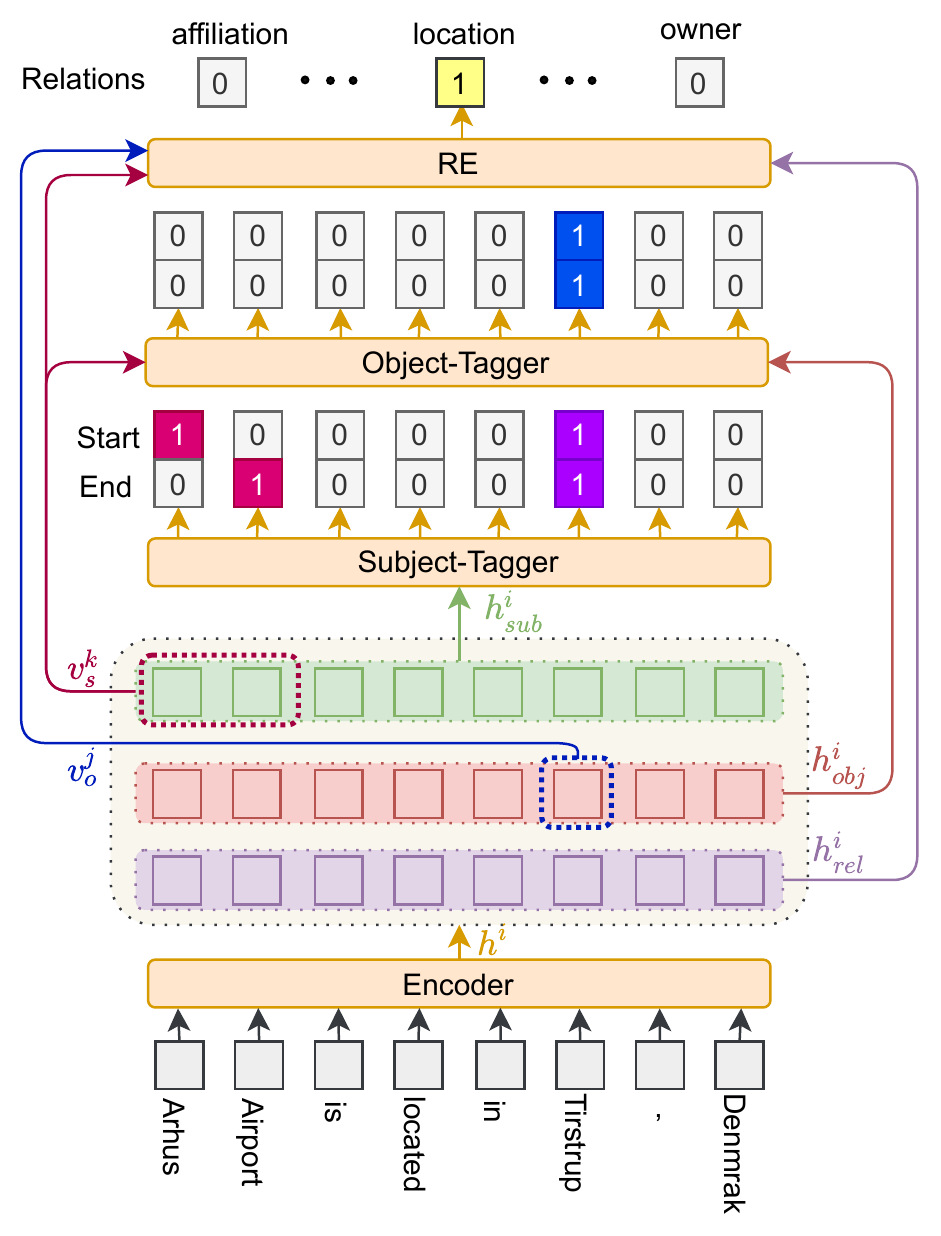}
	\caption{Model Architecture}
	\label{fig:model}
\end{figure}

The architecture of  \emph{ConCasRTE}  is shown in Figure~\ref{fig:model}. There are four main modules in it: an \emph{Encoder} module, a  \emph{Subject-Tagger}  module, an  \emph{Object-Tagger}  module, and a \emph{Relation Extraction} module (\emph{RE} for short). These modules work in a cascade manner. And the latter three modules form a three-step extraction framework: first extracts subjects, then extracts objects, and finally extracts relations.

\noindent\textbf{Encoder}
Firstly, a pre-trained \emph{BERT-Base (Cased)} model ~\cite{devlin2018bert} is used   to generate an initial representation (denoted as $\mathbf{h}^{i} \in \mathbb{R}^{d_h}$) for each token in an input sentence. Then  the context features for subjects, objects, and relations  are generated with Eq.\eqref{eq:2}, where $\mathbf{W}_{(.)} \in \mathbb{R}^{d_h \times d_h}$ are trainable weights, and $\mathbf{b}_{(.)} \in \mathbb{R}^{d_h} $ are biases. 
\begin{equation}
	\begin{aligned}
		&\mathbf{h}_{sub}^i = \mathbf{W}_{sub} \mathbf{h}^i + \mathbf{b}_{sub} \\&\mathbf{h}_{obj}^i = \mathbf{W}_{obj} \mathbf{h}^i + \mathbf{b}_{obj}  \\ &\mathbf{h}_{rel}^i = \mathbf{W}_{rel} \mathbf{h}^i + \mathbf{b}_{rel} \label{eq:2}
	\end{aligned}
\end{equation}



\noindent\textbf{Subject/Object Taggers} Taking each token in a sentence as input, 
\emph{Subject-Tagger} uses two binary  tag sequences to determine whether it   is the start   and end tokens of a subject, as shown  in  Eq.\eqref{eq:ps}.
\begin{equation}
	\begin{aligned}
		&p_{start}^{s,i} = \sigma(\mathbf W_{start}^{s}\mathbf{h}_{sub}^{i}+\mathbf b_{start}^s)\\
		&p_{end}^{s,i} = \sigma(\mathbf W_{end}^{s}\mathbf{h}_{sub}^{i}+\mathbf b_{end}^s) \label{eq:ps}
	\end{aligned}
\end{equation}
where $p_{start}^{s,i}$ and $p_{end}^{s,i}$ denote the probabilities of the $i$-th input token  being the start   and end tokens of a subject respectively. 

Subsequently, taking each extracted subject    as an input prior condition, \emph{Object-Tagger} extracts all objects of this subject. It also uses two binary tag sequences to determine  whether a token in the input sentence is the   start  and  end tokens of an object that can form a (\emph{subject, object}) pair with the input subject, as  shown in Eq.\eqref{eq:po}.
\begin{equation}
	\begin{aligned}
		&{p}_{start}^{o,i,k} = \sigma(\mathbf W_{start}^{o}(\mathbf{h}_{obj}^{i} \circ \mathbf{v}_{s}^{k})+\mathbf b_{start}^o)\\
		&{p}_{end}^{o,i,k} = \sigma(\mathbf W_{end}^{o}(\mathbf{h}_{obj}^{i} \circ \mathbf{v}_{s}^{k})+\mathbf b_{end}^o) \label{eq:po}
	\end{aligned}
\end{equation}
where $\mathbf{v}_s^k$ is the vector representation of the $k$-th input subject and  is obtained by simply averaging all its tokens' vector representations;   $p_{start}^{o,i,k}$ and $p_{end}^{o,i,k}$ denote the probabilities of the $i$-th input token  being the start and end tokens of an object that can form an entity pair with the $k$-th input {subject};  $\circ$ denotes the hadamard product operation.   




\noindent\textbf{RE}
Taking each (\emph{subject, object}) pair as input,  \emph{RE}   extracts  all   relations for this input entity pair, as   shown in Eq.~\eqref{eq:pr}. 
\begin{equation}
	\begin{aligned}
		&\mathbf{p}_{r}^{k,j} = \frac{1}{\vert LOC \vert}\sum_{i \in \emph{LOC}} \sigma\left(\mathbf{W}_{r}\left(\mathbf{h}_{rel}^{i} \circ \mathbf{v}_{s}^{k} \circ \mathbf{v}_{o}^{j}\right)+ \mathbf{b}_{r}^{k,j}\right)\\ 
		&\emph{LOC} = [{loc}_{s}^{k, start}, {loc}_{s}^{k, end} ] \cup [{loc}_{o}^{j, start}, {loc}_{o}^{j, end}]  \label{eq:pr}
	\end{aligned}
\end{equation}
where $\mathbf{v}_s^k$ and $\mathbf{v}_o^j$ are   vector representations of the $k$-th subject and $j$-th object, and $\mathbf{v}_o^j$  is obtained by the same way as $\mathbf{v}_s^k$; $ \mathbf{p}_{r}^{k,j} \in \mathbb{R}^{\vert \emph{R} \vert}$ is a probability sequence, ${\vert \emph{R} \vert}$ is the size of  relation set  $\emph{R}$, and each   item in   $ \mathbf{p}_{r}^{k,j}$ corresponds to a specific relation and is used to determine whether this relation should be assigned to the input entity  pair;   
${loc}_{s}^{k, start}$, ${loc}_{s}^{k, end}$, ${loc}_{o}^{j, start}$ and ${loc}_{o}^{j, end}$ denote the start and end positions of the two input entities; $\emph{LOC}$ is the position range of the input entity pair, and  $|LOC|$ is the number of tokens in this pair. 

In Eq.\eqref{eq:ps}-\eqref{eq:pr}, $\mathbf{W}_{(.)}^s,\mathbf{W}_{\left(.\right)}^o \in \mathbb{R}^{1 \times d_h}, \mathbf{W}_{r} \in \mathbb{R}^{\vert \emph{R} \vert \times d_h}$  are  weights,  $\mathbf{b}_{(.)}^s, \mathbf{b}_{\left(.\right)}^o\in \mathbb{R}^1, \mathbf{b}_{r}^{(,)} \in \mathbb{R}^{\vert \emph{R} \vert} $ are biases,  and $\sigma$ is a sigmoid  function. 



\noindent\textbf{\emph{Confidence Threshold} based Loss} Traditional loss functions like cross entropy usually assign lower costs to samples whose predictions are  correct and the model is   \emph{confident}  for these predictions (here we say a model is \emph{confident} for a prediction if it assigns a very high or a very low probability for this prediction).  The major classes  usually account  for the majority of these low cost samples due to the  overwhelming number of samples in them.  
So it would bring following  two benefits if we directly neglect  these low cost samples.  First, most of the neglected samples would be in the major classes. Second, neglecting these samples wouldn't have much impact on  the  model training since the predictions of these  samples are correct and \emph{confident}.  Accordingly,  the \emph{class imbalance issue} would be alleviated greatly by such  a neglect operation. 
Inspired by this, we propose a \emph{confidence threshold} based cross entropy loss which makes a model only be trained by the samples whose predictions are not \emph{confident} or incorrect, as shown in  Eq.~\eqref{eq:bce2}-~\eqref{eq:lossWeight}. 
\begin{align}
	&ce^{\prime}\left(p, t\right)=\xi * {ce }\left(p , t\right) \label{eq:bce2} \\
	& ce(p,t)= -[t log p+(1-t)  log(1-p)] \label{eq:loss} \\
	& \xi=\begin{cases}
		0,\left(t-T\right) \left(p-T\right)>0 \& \left|p-0.5\right|>C \\
		1,  otherwise
	\end{cases} \label{eq:lossWeight}
\end{align}
where $ce^{\prime}$ is the proposed  loss; $ce$ is a basic binary cross entropy loss; $p \in (0,1)$ is a prediction probability and $t \in \{0, 1\}$ is its true tag;  $\xi$ is a switch  coefficient to determine whether the model be trained by an input sample; $T$ is a hyperparameter used to determine whether a   prediction  is assigned  $1$ or $0$;  $C\in[0,0.5]$ is a hyperparameter and we call it as  \emph{confidence threshold}; $|p-0.5|>C$ means the model is \emph{confident} for the  prediction: the larger  the \emph{confidence threshold} is set, the higher confident degree of the model for its predictions is required; and    $\left(t-T\right) \left(p-T\right)>0$ means  the prediction is correct.




Finally, the proposed loss is used   for training  the modules of \emph{Subject-Tagger}, \emph{Object-Tagger}, and \emph{RE}. The   overall loss  of \emph{ConCasRTE}  is defined as the sum of these separated  losses. During training, we take the popular \emph{teacher forcing} strategy where the ground truth samples are used as input. To alleviate the \emph{exposure bias} issue~\cite{wang-etal-2020-tplinker} caused by this strategy,  we  add  some randomly generated noise samples into the ground truth samples and use them together. 

\section{Experiments}

\label{section:Experiment}
\subsection{Experiment Settings}
\noindent \textbf{Datasets}  Here following two benchmark datasets are used: NYT~\cite{riedel2010Modeling} and WebNLG~\cite{gardent2017Creating}. Both of them have two different versions according to following two annotation  standards: 1) annotating the last token of each entity, and 2) annotating the whole entity span. Following \emph{TPLinker}~\cite{wang-etal-2020-tplinker}, we denote the datasets based on the first standard as NYT$^*$ and WebNLG$^*$, and the datasets based on the second  standard as NYT and WebNLG. Some  statistics of these  datasets are shown in Table~\ref{tab:statistics}: \emph{EPO}, \emph{SEO}, and \emph{Normal} refer to \emph{entity pair overlapping}, \emph{single entity overlapping}, and \emph{no overlapped triples} respectively~\cite{zeng2018Extracting}. Note a sentence can belong to both \emph{EPO}  and \emph{SEO}.

\noindent \textbf{Evaluation Metrics} The standard micro precision, recall, and \emph{F1} score are used to evaluate the results. 
There are two match standards for the RTE task:  (i) \emph{Partial Match}: an extracted triplet is regarded as correct if the predicted relation and the head of both subject entity and object entity are  correct; (ii) \emph{Exact Match}: a  triple is regarded as  correct  only when its  entities and relation are completely matched with a correct triple. Here we follow ~\cite{wang-etal-2020-tplinker,wei-etal-2020-novel,sun2021}: use \emph{Partial Match} on NYT$^*$ and WebNLG$^*$, and use \emph{Exact Match} on NYT and WebNLG. 


\noindent \textbf{Implementation Details} 
AdamW~\cite{KingmaB14} is used to train \emph{ConCasRTE}. All the hyperparameters are determined based on the results on the development set. Finally,   they are  set as follows. On  NYT and NYT$^*$, the  batch size is set to 18 and epoch is set to 100. On WebNLG and WebNLG$^*$, the  batch size is set to 6 and epoch is set to 50. On all datasets, the learning rate is set to $1e^{-5}$,  the   \emph{confidence threshold} ($C$ in Eq.~\eqref{eq:lossWeight})  is set to $0.1$, and all other   thresholds   are  set to $0.5$.  

\noindent \textbf{Baselines} Following  strong state-of-the-art  models are taken as baselines: 
\emph{ETL-Span}~\cite{yu2019joint}, \emph{WDec}~\cite{nayak2020effective},
\emph{RSAN}~\cite{yuan2020a}, \emph{RIN}~\cite{SunZMML20}, \emph{PMEI}~\cite{sun2021}, \emph{CasRel}~\cite{wei-etal-2020-novel}, and  \emph{TPLinker}~\cite{wang-etal-2020-tplinker}.  
We also implement a \emph{LSTM}-encoder version of \emph{ConCasRTE}  where   300-dimensional GloVe embeddings~\cite{pennington2014Glove} and 2-layer stacked BiLSTM are used. 

\subsection{Experimental Results}
\noindent \textbf{Main Results} The main experimental results are shown in Table ~\ref{table:main}. We can see  that  \emph{ConCasRTE} is very effective. On all datasets and under both match standards, it consistently outperforms all the compared state-of-the-art baselines  in term of \emph{F1}. As for other metrics,  \emph{ConCasRTE}  achieves the best results on most of cases, and even the  exceptions are very close to the best results. 
\begin{table}[t] 
	\caption{Statistics of datasets.} 
	\label{tab:statistics} 
	\centering 
	\begin{tabular}{ccccc} 
		\hline
		\multirow{2}{*}{Category}&
		\multicolumn{2}{c}{NYT}&
		\multicolumn{2}{c}{WebNLG}
		
		\\
		\cmidrule(lr){2-3}\cmidrule(lr){4-5}
		&Train&Test&Train&Test\\
		\hline 
		\textit{Normal}&37013&3266&1596&246 \\
		\textit{EPO}&9782&978&227&26 \\
		\textit{SEO}&14735&1297&3406&457\\
		\hline 
	\end{tabular} 
	
\end{table}

\noindent\textbf{{Evaluations on Complex Sentences}} Here we  evaluate \emph{ConCasRTE}'s  ability for extracting triples from complex sentences that contain overlapping triples or multiple triples.  This ability is widely  discussed by existing  work, and can be viewed as an important metric to evaluate the robustness of a model. For fair comparison,    we follow the  settings of some previous best models~\cite{wei-etal-2020-novel,wang-etal-2020-tplinker}: (i) classifying sentences according to the degree of entity overlapping and the number of triples contained in a sentence, and (ii) conducting experiments on different subsets of NYT$^*$ and WebNLG$^*$.



The results are  in Table~\ref{table:f1_on_split}, which demonstrate  the great  superiority of \emph{ConCasRTE}  for handling both kinds of complex sentences. On both datasets, it  achieves   much better results than the compared baselines. 
In fact,  \emph{ConCasRTE} inherits the main strengths of existing tagging based methods for extracting triples from complex sentences, while well addresses the \emph{class imbalance issue} existed in  these methods, thus it achieves much better results. 

\begin{table*}[htbp]
	\caption{Main experiments.  $^\star$ means the results are  produced by us by running  the available source code.  }
	\label{table:main}
	\begin{center}
		
		{\begin{tabular}{lccccccccccccc}
				\hline
				\multirow{3}{*}{Model}   
				&\multicolumn{6}{c}{Partial Match} & \multicolumn{6}{c}{Exact Match}& \\
				\cmidrule(lr){2-7} 
				\cmidrule(lr){8-13} 
				&\multicolumn{3}{c}{NYT$^*$} & \multicolumn{3}{c}{WebNLG$^*$}& \multicolumn{3}{c}{NYT} & \multicolumn{3}{c}{WebNLG}\\
				\cmidrule(lr){2-4} 
				\cmidrule(lr){5-7} 
				\cmidrule(lr){8-10} 
				\cmidrule(lr){11-13} 
				& Prec. & Rec. & F1 & Prec. & Rec. & F1 & Prec. & Rec. & F1 & Prec. & Rec. & F1 \\
				\hline
				ETL-Span &84.9 &72.3 &78.1 &84.0 &91.5 &87.6 &85.5 &71.7 &78.0 &84.3 &82.0 &83.1\\
				WDec &-- &-- &-- &-- &-- &-- &88.1 &76.1 &81.7 &-- &-- &--\\
				RSAN &-- &-- &-- &-- &-- &-- &85.7 &83.6 &84.6 &80.5 &83.8 &82.1\\
				RIN &87.2 &87.3 &87.3 &87.6 &87.0 &87.3 &83.9 &85.5 &84.7 &77.3 &76.8 &77.0\\
				CasRel$_{LSTM}$ &84.2 &83.0 &83.6 &86.9 &80.6 &83.7 &-- &-- &-- &-- &-- &--\\
				PMEI$_{LSTM}$ &88.7 &86.8 &87.8 &88.7 &87.6 &88.1 &84.5 &84.0 &84.2 &78.8 &77.7 &78.2\\
				TPLinker$_{LSTM}$ &83.8 &83.4 &83.6 &90.8 &90.3 &90.5 &86.0 &82.0 &84.0 &\textbf{91.9} &81.6 &86.4\\
				
				CasRel$_{BERT}$ &89.7 &89.5 &89.6 &93.4 &90.1 &91.8 &89.8$^{\star}$ &88.2$^{\star}$ &89.0$^{\star}$  &88.3$^{\star}$ &84.6$^{\star}$ &86.4$^{\star}$\\
				PMEI$_{BERT}$ &90.5 &89.8 &90.1 &91.0 &92.9 &92.0 &88.4 &88.9 &88.7 &80.8 &82.8 &81.8\\
				TPLinker$_{BERT}$ &91.3 &\textbf{92.5} &91.9 &91.8 &92.0 &91.9 &91.4 &\textbf{92.6} &92.0 &88.9 &84.5 &86.7\\		
				\hline
				\textbf{ConCasRTE}$_{LSTM}$ &88.1&86.6&87.3 &91.2&90.8&91.0 &86.6&82.3&84.4 &88.3&83.9&86.0 \\
				\textbf{ConCasRTE}$_{BERT}$ &\textbf{92.9}&92.3&\textbf{92.6} &\textbf{93.8}&\textbf{92.5}&\textbf{93.1} &\textbf{92.9}&92.1&\textbf{92.5}&90.6&\textbf{88.1}&\textbf{89.3} \\
				\hline
		\end{tabular}}
	\end{center}
\end{table*}
\begin{table*}[t]
	\caption{F1 scores on sentences with different overlapping pattern and different triplet number. Results of \emph{CasRel} are copied from \emph{TPLinker} directly. ``T'' is the number of triples contained in a sentence. 
	}
	\label{table:f1_on_split}
	\begin{center}
		
		\setlength{\tabcolsep}{0.5 mm}{\begin{tabular}{lccccccccccccccccc}
				\hline
				\multirow{2}{*}{Model} & \multicolumn{8}{c}{NYT$^*$}& \multicolumn{9}{c}{WebNLG$^*$} \\
				& Normal & SEO & EPO & T = 1 & T = 2 & T = 3 & T = 4 & T $\geq$ 5 && Normal & SEO & EPO & T = 1 & T = 2 & T = 3 & T = 4 & T $\geq$ 5 \\
				\hline
				CasRel$_{BERT}$  & 87.3 & 91.4 & 92.0 & 88.2 & 90.3 & 91.9 & 94.2 & 83.7 &&89.4 & 92.2 & 94.7 &89.3 &90.8 & 94.2 & 92.4 & 90.9  \\
				TPLinker$_{BERT}$  & {90.1} & {93.4} & {94.0} & {90.0} & {92.8} & {93.1} & \textbf{96.1} & {90.0} && 87.9 & {92.5} & \textbf{95.3} & 88.0 & 90.1 & {94.6} & {93.3} & {91.6}\\ 
				\hline
				ConCasRTE$_{BERT}$ &\textbf{90.6} &\textbf{94.0} &\textbf{94.1} &\textbf{90.5} &\textbf{93.8} &\textbf{93.4} &95.2 &\textbf{91.7} & &\textbf{91.1} &\textbf{93.3} &93.8 &\textbf{90.7} &\textbf{91.9} &\textbf{95.5} &\textbf{93.4} &\textbf{91.9}\\
				\hline
		\end{tabular}}
		
	\end{center}
\end{table*}
\begin{table}[htbp]
	\caption{Detailed Results  (F1). $\uparrow$ means increased scores.}
	\label{tab:loss}
	\centering
	{
		\begin{tabular}{lllll}
			\hline
			\multirow{1}{*}{Models}&\multicolumn{1}{l}{NYT$^*$}&\multicolumn{1}{l}{WebNLG$^*$}&\multicolumn{1}{l}{NYT}&\multicolumn{1}{l}{WebNLG}\\
			\midrule

			
			\multirow{1}{*}ConCasRTE$_{Ce}$ &91.8&91.9&91.6&87.9\\
			\hline
			\multirow{1}{*}ConCasRTE$_{DifW}$ &92.1&92.4&91.8&88.4\\
			\multirow{1}{*}ConCasRTE$_{ReS}$ &92.5&92.7&91.9&88.7\\
			\multirow{1}{*}ConCasRTE$_{FLos}$ &92.5&92.9&92.2&89.0\\
			\hline
			
			\multirow{1}{*}{ETL-Span$_{CLos}$} &78.9($\uparrow$0.8) &88.8($\uparrow$1.2)&78.8($\uparrow$0.8)&84.1($\uparrow$1.0)\\
			\multirow{1}{*}{CasRel$_{CLos}$} &90.0($\uparrow$0.4) &92.3($\uparrow$0.5)&89.6($\uparrow$0.6)&87.9($\uparrow$1.5)\\
			\multirow{1}{*}{TPLinker$_{CLos}$} &92.2($\uparrow$0.3) &92.5($\uparrow$0.6)&92.3($\uparrow$0.3)&88.1($\uparrow$1.4)\\
			\hline
		\end{tabular}
	}
	
\end{table}

\noindent\textbf{Detailed Analyses}   Table~\ref{tab:loss}   shows some detailed experimental results about  the proposed  extraction framework and   loss function. All  these results are obtained when the \emph{BERT}-based encoder used.

\textbf{First}, we evaluate the effectiveness of the proposed extraction framework. To this end, we implement {ConCasRTE}$_{Ce}$, a variant  that uses the basic binary cross entropy  loss. Then we compare it with  \emph{CasRel} (the current best tagging based RTE model) since  the main  difference between them is the extraction framework. We can see    {ConCasRTE}$_{Ce}$  achieves much  better results on all datasets. In fact, the proposed  framework can reduce the total number of samples greatly, which is much helpful for alleviating the \emph{class imbalance issue}.  Taking a $l$-token sentence as example,  the number of samples  in \emph{ConCasRTE} is $2l+ 2sl+n|R|$  ($s$ is the number of  subjects extracted, and  $n$ is the number of all (\emph{subject, object}) pairs). In this number, $2l$, $2sl$, and  $n|R|$ are generated by the modules of \emph{Subject-Tagger},  \emph{Object-Tagger}, and  \emph{RE} respectively.  In  \emph{CasRel}, the number of samples   is $2l+2sl|R|$, where  $2l$ and $2sl|R|$ are generated by its modules of subject extraction  and object-relation extraction  respectively.   Usually  $n \ll l$, thus   $2l+ 2sl+n|R|\ll 2l+ 2sl+l|R| < 2l+2sl|R|$. And there are $2s+2n+t$ and $2s+2t$ samples in the positive classes of \emph{ConCasRTE} and \emph{CasRel}   respectively ($t$ is the number of  triples), and the difference between these two numbers can be negligible since  both are very small. 
So  the number gap between samples in classes of positive and negative  in \emph{ConCasRTE} is much  smaller than that in \emph{CasRel}, which makes the \emph{class imbalance issue}  alleviated greatly. 

\textbf{Second}, we evaluate  the proposed loss function from the aspects of ability for addressing the \emph{class imbalance issue} and   adaptability. 

(i) \emph{Ability Evaluation}.  To evaluate the proposed loss function's ability   for addressing the \emph{class imbalance issue}, we implement following variants that use different methods for addressing the mentioned issue. (1) ConCasRTE$_{DifW}$:  a variant that assigns different weights for  the losses of positive and negative classes (here  0.75 for the positive  and 0.25 for the negative). (2) ConCasRTE$_{ReS}$: a re-sampling based variant that randomly selects some samples from the negative class so that makes the proportion between samples in the classes of positive and negative be a predefined threshold (here is 1:5). (3) ConCasRTE$_{FLoss}$, a variant that uses  \emph{Focal Loss}~\cite{lin2020focal}  (its hyperparameter  $\gamma$  is set to $2$). 
We can see that the proposed loss brings  the greatest performance   improvement over ConCasRTE$_{Ce}$ than all the compared methods, which demonstrates the proposed loss  is more effective.  Different from existing state-of-the-art methods like  \emph{Focal Loss}, the proposed loss function does not try to increase the importance of  samples in the minor classes. Instead, it   directly removes some samples in the major classes so as to   narrow  the number gap between samples in the major and minor classes. These comparison results show  this strategy  is  more effective. 


(ii) \emph{Adaptability Evaluation}. The proposed loss is applicable to a wide range of models since we don't make any assumptions about the data distribution. For example, it can be used not only in the tagging based methods, but also in other kinds of methods. To evaluate this, we transplant it to following diverse  models including  \emph{ETL-Span}, \emph{CasRel} ,  and \emph{TPLinker}. These new models are marked by a subscript ``\emph{CLos}''.  Results show that all these new models  achieve significant improvement over   their original ones on  all  datasets. 

\section{Conclusions}
In this paper, we propose a novel conditional cascade  RTE model. It  contains following two novelties for  addressing the \emph{class imbalance issue}  existed in the tagging based methods. First,  we propose a simple but effective  three-step  extraction framework. Second, we propose  an effective and adaptive \emph{confidence threshold} based cross entropy  loss function. We evaluate the proposed model on two benchmark datasets. Experiments show that both novelties can alleviate the \emph{class imbalance issue} effectively, and they help the proposed model achieve  state-of-the-art results on both datasets. 

\begin{acks}
	This work is supported by the National Key R\&D Program of China (No.2018YFC0830701), the National Natural Science Foundation of China (No.61572120), the Fundamental Research Funds for the Central Universities (No.N181602013 and No.N171602003), Ten Thousand Talent Program (No.ZX20200035), and Liaoning Distinguished Professor (No.XLYC1902057).
\end{acks}
\bibliographystyle{ACM-Reference-Format}
\bibliography{sample-base}


\begin{thebibliography}{31}


\ifx \showCODEN    \undefined \def \showCODEN     #1{\unskip}     \fi
\ifx \showDOI      \undefined \def \showDOI       #1{#1}\fi
\ifx \showISBNx    \undefined \def \showISBNx     #1{\unskip}     \fi
\ifx \showISBNxiii \undefined \def \showISBNxiii  #1{\unskip}     \fi
\ifx \showISSN     \undefined \def \showISSN      #1{\unskip}     \fi
\ifx \showLCCN     \undefined \def \showLCCN      #1{\unskip}     \fi
\ifx \shownote     \undefined \def \shownote      #1{#1}          \fi
\ifx \showarticletitle \undefined \def \showarticletitle #1{#1}   \fi
\ifx \showURL      \undefined \def \showURL       {\relax}        \fi
\providecommand\bibfield[2]{#2}
\providecommand\bibinfo[2]{#2}
\providecommand\natexlab[1]{#1}
\providecommand\showeprint[2][]{arXiv:#2}

\bibitem[\protect\citeauthoryear{{Bekoulis}, {Deleu}, {Demeester}, and
  {Develder}}{{Bekoulis} et~al\mbox{.}}{2018}]%
        {bekoulis2018joint}
\bibfield{author}{\bibinfo{person}{Giannis {Bekoulis}},
  \bibinfo{person}{Johannes {Deleu}}, \bibinfo{person}{Thomas {Demeester}},
  {and} \bibinfo{person}{Chris {Develder}}.} \bibinfo{year}{2018}\natexlab{}.
\newblock \showarticletitle{Joint entity recognition and relation extraction as
  a multi-head selection problem}.
\newblock \bibinfo{journal}{\emph{Expert Systems With Applications}}
  \bibinfo{volume}{114} (\bibinfo{year}{2018}), \bibinfo{pages}{34--45}.
\newblock


\bibitem[\protect\citeauthoryear{{Chan} and {Roth}}{{Chan} and {Roth}}{2011}]%
        {chan2011exploiting}
\bibfield{author}{\bibinfo{person}{Yee~Seng {Chan}} {and} \bibinfo{person}{Dan
  {Roth}}.} \bibinfo{year}{2011}\natexlab{}.
\newblock \showarticletitle{Exploiting Syntactico-Semantic Structures for
  Relation Extraction}. In \bibinfo{booktitle}{\emph{Proceedings of the 49th
  Annual Meeting of the Association for Computational Linguistics: Human
  Language Technologies}}. \bibinfo{pages}{551--560}.
\newblock


\bibitem[\protect\citeauthoryear{{Cui}, {Jia}, {Lin}, {Song}, and
  {Belongie}}{{Cui} et~al\mbox{.}}{2019}]%
        {cui2019class}
\bibfield{author}{\bibinfo{person}{Yin {Cui}}, \bibinfo{person}{Menglin {Jia}},
  \bibinfo{person}{Tsung-Yi {Lin}}, \bibinfo{person}{Yang {Song}}, {and}
  \bibinfo{person}{Serge {Belongie}}.} \bibinfo{year}{2019}\natexlab{}.
\newblock \showarticletitle{Class-Balanced Loss Based on Effective Number of
  Samples}. In \bibinfo{booktitle}{\emph{2019 IEEE/CVF Conference on Computer
  Vision and Pattern Recognition (CVPR)}}. \bibinfo{pages}{9268--9277}.
\newblock


\bibitem[\protect\citeauthoryear{{Devlin}, {Chang}, {Lee}, and
  {Toutanova}}{{Devlin} et~al\mbox{.}}{2018}]%
        {devlin2018bert}
\bibfield{author}{\bibinfo{person}{Jacob {Devlin}}, \bibinfo{person}{Ming-Wei
  {Chang}}, \bibinfo{person}{Kenton {Lee}}, {and} \bibinfo{person}{Kristina~N.
  {Toutanova}}.} \bibinfo{year}{2018}\natexlab{}.
\newblock \showarticletitle{BERT: Pre-training of Deep Bidirectional
  Transformers for Language Understanding}. In
  \bibinfo{booktitle}{\emph{Proceedings of the 2019 Conference of the North
  American Chapter of the Association for Computational Linguistics: Human
  Language Technologies, Volume 1 (Long and Short Papers)}}.
  \bibinfo{pages}{4171--4186}.
\newblock


\bibitem[\protect\citeauthoryear{{Eberts} and {Ulges}}{{Eberts} and
  {Ulges}}{2019}]%
        {eberts2019span}
\bibfield{author}{\bibinfo{person}{Markus {Eberts}} {and}
  \bibinfo{person}{Adrian {Ulges}}.} \bibinfo{year}{2019}\natexlab{}.
\newblock \showarticletitle{Span-Based Joint Entity and Relation Extraction
  with Transformer Pre-Training.}. In \bibinfo{booktitle}{\emph{ECAI}}.
  \bibinfo{pages}{2006--2013}.
\newblock


\bibitem[\protect\citeauthoryear{{Fu}, {Li}, and {Ma}}{{Fu}
  et~al\mbox{.}}{2019}]%
        {fu2019graphrel}
\bibfield{author}{\bibinfo{person}{Tsu-Jui {Fu}}, \bibinfo{person}{Peng-Hsuan
  {Li}}, {and} \bibinfo{person}{Wei-Yun {Ma}}.}
  \bibinfo{year}{2019}\natexlab{}.
\newblock \showarticletitle{GraphRel: Modeling Text as Relational Graphs for
  Joint Entity and Relation Extraction}. In
  \bibinfo{booktitle}{\emph{Proceedings of the 57th Annual Meeting of the
  Association for Computational Linguistics}}. \bibinfo{pages}{1409--1418}.
\newblock


\bibitem[\protect\citeauthoryear{Gardent, Shimorina, Narayan, and
  Perez-Beltrachini}{Gardent et~al\mbox{.}}{2017}]%
        {gardent2017Creating}
\bibfield{author}{\bibinfo{person}{Claire Gardent}, \bibinfo{person}{Anastasia
  Shimorina}, \bibinfo{person}{Shashi Narayan}, {and} \bibinfo{person}{Laura
  Perez-Beltrachini}.} \bibinfo{year}{2017}\natexlab{}.
\newblock \showarticletitle{Creating Training Corpora for NLG Micro-Planners}.
  In \bibinfo{booktitle}{\emph{Proceedings of the 55th Annual Meeting of the
  Association for Computational Linguistics (Volume 1: Long Papers)}}.
  \bibinfo{pages}{179--188}.
\newblock


\bibitem[\protect\citeauthoryear{Gupta, Sch{\"u}tze, and Andrassy}{Gupta
  et~al\mbox{.}}{2016}]%
        {gupta-etal-2016-table}
\bibfield{author}{\bibinfo{person}{Pankaj Gupta}, \bibinfo{person}{Hinrich
  Sch{\"u}tze}, {and} \bibinfo{person}{Bernt Andrassy}.}
  \bibinfo{year}{2016}\natexlab{}.
\newblock \showarticletitle{Table Filling Multi-Task Recurrent Neural Network
  for Joint Entity and Relation Extraction}. In
  \bibinfo{booktitle}{\emph{Proceedings of {COLING} 2016, the 26th
  International Conference on Computational Linguistics: Technical Papers}}.
  \bibinfo{publisher}{The COLING 2016 Organizing Committee},
  \bibinfo{address}{Osaka, Japan}, \bibinfo{pages}{2537--2547}.
\newblock


\bibitem[\protect\citeauthoryear{{Johnson} and {Khoshgoftaar}}{{Johnson} and
  {Khoshgoftaar}}{2019}]%
        {johnson2019survey}
\bibfield{author}{\bibinfo{person}{Justin~M. {Johnson}} {and}
  \bibinfo{person}{Taghi~M. {Khoshgoftaar}}.} \bibinfo{year}{2019}\natexlab{}.
\newblock \showarticletitle{Survey on deep learning with class imbalance}.
\newblock \bibinfo{journal}{\emph{Journal of Big Data}} \bibinfo{volume}{6},
  \bibinfo{number}{1} (\bibinfo{year}{2019}), \bibinfo{pages}{1--54}.
\newblock


\bibitem[\protect\citeauthoryear{Kingma and Ba}{Kingma and Ba}{2015}]%
        {KingmaB14}
\bibfield{author}{\bibinfo{person}{Diederik~P. Kingma} {and}
  \bibinfo{person}{Jimmy Ba}.} \bibinfo{year}{2015}\natexlab{}.
\newblock \showarticletitle{Adam: {A} Method for Stochastic Optimization}. In
  \bibinfo{booktitle}{\emph{3rd International Conference on Learning
  Representations, {ICLR} 2015, San Diego, CA, USA, May 7-9, 2015, Conference
  Track Proceedings}}, \bibfield{editor}{\bibinfo{person}{Yoshua Bengio} {and}
  \bibinfo{person}{Yann LeCun}} (Eds.).
\newblock


\bibitem[\protect\citeauthoryear{{Lin}, {Goyal}, {Girshick}, {He}, and
  {Dollar}}{{Lin} et~al\mbox{.}}{2020}]%
        {lin2020focal}
\bibfield{author}{\bibinfo{person}{Tsung-Yi {Lin}}, \bibinfo{person}{Priya
  {Goyal}}, \bibinfo{person}{Ross {Girshick}}, \bibinfo{person}{Kaiming {He}},
  {and} \bibinfo{person}{Piotr {Dollar}}.} \bibinfo{year}{2020}\natexlab{}.
\newblock \showarticletitle{Focal Loss for Dense Object Detection}.
\newblock \bibinfo{journal}{\emph{IEEE Transactions on Pattern Analysis and
  Machine Intelligence}} \bibinfo{volume}{42}, \bibinfo{number}{2}
  (\bibinfo{year}{2020}), \bibinfo{pages}{318--327}.
\newblock


\bibitem[\protect\citeauthoryear{{Menon}, {Jayasumana}, {Rawat}, {Jain},
  {Veit}, and {Kumar}}{{Menon} et~al\mbox{.}}{2020}]%
        {menon2020long}
\bibfield{author}{\bibinfo{person}{Aditya~Krishna {Menon}},
  \bibinfo{person}{Sadeep {Jayasumana}}, \bibinfo{person}{Ankit~Singh {Rawat}},
  \bibinfo{person}{Himanshu {Jain}}, \bibinfo{person}{Andreas {Veit}}, {and}
  \bibinfo{person}{Sanjiv {Kumar}}.} \bibinfo{year}{2020}\natexlab{}.
\newblock \showarticletitle{Long-tail learning via logit adjustment.}
\newblock \bibinfo{journal}{\emph{arXiv preprint arXiv:2007.07314}}
  (\bibinfo{year}{2020}).
\newblock


\bibitem[\protect\citeauthoryear{Miwa and Bansal}{Miwa and Bansal}{2016}]%
        {miwa-bansal-2016-end}
\bibfield{author}{\bibinfo{person}{Makoto Miwa} {and} \bibinfo{person}{Mohit
  Bansal}.} \bibinfo{year}{2016}\natexlab{}.
\newblock \showarticletitle{End-to-End Relation Extraction using {LSTM}s on
  Sequences and Tree Structures}. In \bibinfo{booktitle}{\emph{Proceedings of
  the 54th Annual Meeting of the Association for Computational Linguistics
  (Volume 1: Long Papers)}}. \bibinfo{publisher}{Association for Computational
  Linguistics}, \bibinfo{address}{Berlin, Germany},
  \bibinfo{pages}{1105--1116}.
\newblock


\bibitem[\protect\citeauthoryear{{Nayak} and {Ng}}{{Nayak} and {Ng}}{2020}]%
        {nayak2020effective}
\bibfield{author}{\bibinfo{person}{Tapas {Nayak}} {and}
  \bibinfo{person}{Hwee~Tou {Ng}}.} \bibinfo{year}{2020}\natexlab{}.
\newblock \showarticletitle{Effective Modeling of Encoder-Decoder Architecture
  for Joint Entity and Relation Extraction}.
\newblock \bibinfo{journal}{\emph{Proceedings of the AAAI Conference on
  Artificial Intelligence}} \bibinfo{volume}{34}, \bibinfo{number}{5}
  (\bibinfo{year}{2020}), \bibinfo{pages}{8528--8535}.
\newblock


\bibitem[\protect\citeauthoryear{Nayak and Ng}{Nayak and Ng}{2020}]%
        {nayakn20}
\bibfield{author}{\bibinfo{person}{Tapas Nayak} {and} \bibinfo{person}{Hwee~Tou
  Ng}.} \bibinfo{year}{2020}\natexlab{}.
\newblock \showarticletitle{Effective Modeling of Encoder-Decoder Architecture
  for Joint Entity and Relation Extraction}. In \bibinfo{booktitle}{\emph{The
  Thirty-Fourth {AAAI} Conference on Artificial Intelligence, {AAAI} 2020, The
  Thirty-Second Innovative Applications of Artificial Intelligence Conference,
  {IAAI} 2020, The Tenth {AAAI} Symposium on Educational Advances in Artificial
  Intelligence, {EAAI} 2020, New York, NY, USA, February 7-12, 2020}}.
  \bibinfo{publisher}{{AAAI} Press}, \bibinfo{pages}{8528--8535}.
\newblock


\bibitem[\protect\citeauthoryear{Pennington, Socher, and Manning}{Pennington
  et~al\mbox{.}}{2014}]%
        {pennington2014Glove}
\bibfield{author}{\bibinfo{person}{Jeffrey Pennington},
  \bibinfo{person}{Richard Socher}, {and} \bibinfo{person}{Christopher
  Manning}.} \bibinfo{year}{2014}\natexlab{}.
\newblock \showarticletitle{Glove: Global vectors for word representation}. In
  \bibinfo{booktitle}{\emph{Proceedings of the 2014 Conference on Empirical
  Methods in Natural Language Processing (EMNLP)}}.
  \bibinfo{pages}{1532--1543}.
\newblock


\bibitem[\protect\citeauthoryear{Riedel, Yao, and McCallum}{Riedel
  et~al\mbox{.}}{2010}]%
        {riedel2010Modeling}
\bibfield{author}{\bibinfo{person}{Sebastian Riedel}, \bibinfo{person}{Limin
  Yao}, {and} \bibinfo{person}{Andrew McCallum}.}
  \bibinfo{year}{2010}\natexlab{}.
\newblock \showarticletitle{Modeling relations and their mentions without
  labeled text}. In \bibinfo{booktitle}{\emph{Joint European Conference on
  Machine Learning and Knowledge Discovery in Databases}}.
  \bibinfo{pages}{148--163}.
\newblock


\bibitem[\protect\citeauthoryear{Sun, Zhang, Mensah, Mao, and Liu}{Sun
  et~al\mbox{.}}{2020}]%
        {SunZMML20}
\bibfield{author}{\bibinfo{person}{Kai Sun}, \bibinfo{person}{Richong Zhang},
  \bibinfo{person}{Samuel Mensah}, \bibinfo{person}{Yongyi Mao}, {and}
  \bibinfo{person}{Xudong Liu}.} \bibinfo{year}{2020}\natexlab{}.
\newblock \showarticletitle{Recurrent Interaction Network for Jointly
  Extracting Entities and Classifying Relations}. In
  \bibinfo{booktitle}{\emph{Proceedings of the 2020 Conference on Empirical
  Methods in Natural Language Processing, {EMNLP} 2020, Online, November 16-20,
  2020}}, \bibfield{editor}{\bibinfo{person}{Bonnie Webber},
  \bibinfo{person}{Trevor Cohn}, \bibinfo{person}{Yulan He}, {and}
  \bibinfo{person}{Yang Liu}} (Eds.). \bibinfo{publisher}{Association for
  Computational Linguistics}, \bibinfo{pages}{3722--3732}.
\newblock


\bibitem[\protect\citeauthoryear{Sun, Zhang, Mensah, Mao, and Liu}{Sun
  et~al\mbox{.}}{2021}]%
        {sun2021}
\bibfield{author}{\bibinfo{person}{Kai Sun}, \bibinfo{person}{Richong Zhang},
  \bibinfo{person}{Samuel Mensah}, \bibinfo{person}{Yongyi Mao}, {and}
  \bibinfo{person}{Xudong Liu}.} \bibinfo{year}{2021}\natexlab{}.
\newblock \showarticletitle{Progressive Multitask Learning with Controlled
  Information Flow for Joint Entity and Relation Extraction}. In
  \bibinfo{booktitle}{\emph{Association for the Advancement of Artificial
  Intelligence (AAAI)}}.
\newblock


\bibitem[\protect\citeauthoryear{Wang, Yu, Zhang, Liu, Zhu, and Sun}{Wang
  et~al\mbox{.}}{2020}]%
        {wang-etal-2020-tplinker}
\bibfield{author}{\bibinfo{person}{Yucheng Wang}, \bibinfo{person}{Bowen Yu},
  \bibinfo{person}{Yueyang Zhang}, \bibinfo{person}{Tingwen Liu},
  \bibinfo{person}{Hongsong Zhu}, {and} \bibinfo{person}{Limin Sun}.}
  \bibinfo{year}{2020}\natexlab{}.
\newblock \showarticletitle{{TPL}inker: Single-stage Joint Extraction of
  Entities and Relations Through Token Pair Linking}. In
  \bibinfo{booktitle}{\emph{Proceedings of the 28th International Conference on
  Computational Linguistics}}. \bibinfo{address}{Barcelona, Spain (Online)},
  \bibinfo{pages}{1572--1582}.
\newblock


\bibitem[\protect\citeauthoryear{Wei, Su, Wang, Tian, and Chang}{Wei
  et~al\mbox{.}}{2020}]%
        {wei-etal-2020-novel}
\bibfield{author}{\bibinfo{person}{Zhepei Wei}, \bibinfo{person}{Jianlin Su},
  \bibinfo{person}{Yue Wang}, \bibinfo{person}{Yuan Tian}, {and}
  \bibinfo{person}{Yi Chang}.} \bibinfo{year}{2020}\natexlab{}.
\newblock \showarticletitle{A Novel Cascade Binary Tagging Framework for
  Relational Triple Extraction}. In \bibinfo{booktitle}{\emph{Proceedings of
  the 58th Annual Meeting of the Association for Computational Linguistics}}.
  \bibinfo{publisher}{Association for Computational Linguistics},
  \bibinfo{address}{Online}, \bibinfo{pages}{1476--1488}.
\newblock


\bibitem[\protect\citeauthoryear{{Yu}, {Zhang}, {Shu}, {Liu}, {Wang}, {Wang},
  and {Li}}{{Yu} et~al\mbox{.}}{2019}]%
        {yu2019joint}
\bibfield{author}{\bibinfo{person}{Bowen {Yu}}, \bibinfo{person}{Zhenyu
  {Zhang}}, \bibinfo{person}{Xiaobo {Shu}}, \bibinfo{person}{Tingwen {Liu}},
  \bibinfo{person}{Yubin {Wang}}, \bibinfo{person}{Bin {Wang}}, {and}
  \bibinfo{person}{Sujian {Li}}.} \bibinfo{year}{2019}\natexlab{}.
\newblock \showarticletitle{Joint Extraction of Entities and Relations Based on
  a Novel Decomposition Strategy.}. In \bibinfo{booktitle}{\emph{ECAI}}.
  \bibinfo{pages}{2282--2289}.
\newblock


\bibitem[\protect\citeauthoryear{{Yuan}, {Zhou}, {Pan}, {Zhu}, {Song}, and
  {Guo}}{{Yuan} et~al\mbox{.}}{2020}]%
        {yuan2020a}
\bibfield{author}{\bibinfo{person}{Yue {Yuan}}, \bibinfo{person}{Xiaofei
  {Zhou}}, \bibinfo{person}{Shirui {Pan}}, \bibinfo{person}{Qiannan {Zhu}},
  \bibinfo{person}{Zeliang {Song}}, {and} \bibinfo{person}{Li {Guo}}.}
  \bibinfo{year}{2020}\natexlab{}.
\newblock \showarticletitle{A relation-specific attention network for joint
  entity and relation extraction}. In \bibinfo{booktitle}{\emph{Proceedings of
  the Twenty-Ninth International Joint Conference on Artificial Intelligence}},
  Vol.~\bibinfo{volume}{4}. \bibinfo{pages}{4054--4060}.
\newblock


\bibitem[\protect\citeauthoryear{{Zelenko}, {Aone}, and
  {Richardella}}{{Zelenko} et~al\mbox{.}}{2003}]%
        {zelenko2003kernel}
\bibfield{author}{\bibinfo{person}{Dmitry {Zelenko}}, \bibinfo{person}{Chinatsu
  {Aone}}, {and} \bibinfo{person}{Anthony {Richardella}}.}
  \bibinfo{year}{2003}\natexlab{}.
\newblock \showarticletitle{Kernel methods for relation extraction}.
\newblock \bibinfo{journal}{\emph{Journal of Machine Learning Research}}
  \bibinfo{volume}{3}, \bibinfo{number}{6} (\bibinfo{year}{2003}),
  \bibinfo{pages}{1083--1106}.
\newblock


\bibitem[\protect\citeauthoryear{{Zeng}, {Zhang}, and {Liu}}{{Zeng}
  et~al\mbox{.}}{2020}]%
        {zeng2020copymtl}
\bibfield{author}{\bibinfo{person}{Daojian {Zeng}}, \bibinfo{person}{Haoran
  {Zhang}}, {and} \bibinfo{person}{Qianying {Liu}}.}
  \bibinfo{year}{2020}\natexlab{}.
\newblock \showarticletitle{CopyMTL: Copy Mechanism for Joint Extraction of
  Entities and Relations with Multi-Task Learning}.
\newblock \bibinfo{journal}{\emph{Proceedings of the AAAI Conference on
  Artificial Intelligence}} \bibinfo{volume}{34}, \bibinfo{number}{5}
  (\bibinfo{year}{2020}), \bibinfo{pages}{9507--9514}.
\newblock


\bibitem[\protect\citeauthoryear{{Zeng}, {He}, {Zeng}, {Liu}, {Liu}, and
  {Zhao}}{{Zeng} et~al\mbox{.}}{2019}]%
        {zeng2019Learning}
\bibfield{author}{\bibinfo{person}{Xiangrong {Zeng}}, \bibinfo{person}{Shizhu
  {He}}, \bibinfo{person}{Daojian {Zeng}}, \bibinfo{person}{Kang {Liu}},
  \bibinfo{person}{Shengping {Liu}}, {and} \bibinfo{person}{Jun {Zhao}}.}
  \bibinfo{year}{2019}\natexlab{}.
\newblock \showarticletitle{Learning the Extraction Order of Multiple
  Relational Facts in a Sentence with Reinforcement Learning}. In
  \bibinfo{booktitle}{\emph{Proceedings of the 2019 Conference on Empirical
  Methods in Natural Language Processing and the 9th International Joint
  Conference on Natural Language Processing (EMNLP-IJCNLP)}}.
  \bibinfo{pages}{367--377}.
\newblock


\bibitem[\protect\citeauthoryear{{Zeng}, {Zeng}, {He}, {Liu}, and
  {Zhao}}{{Zeng} et~al\mbox{.}}{2018}]%
        {zeng2018Extracting}
\bibfield{author}{\bibinfo{person}{Xiangrong {Zeng}}, \bibinfo{person}{Daojian
  {Zeng}}, \bibinfo{person}{Shizhu {He}}, \bibinfo{person}{Kang {Liu}}, {and}
  \bibinfo{person}{Jun {Zhao}}.} \bibinfo{year}{2018}\natexlab{}.
\newblock \showarticletitle{Extracting Relational Facts by an End-to-End Neural
  Model with Copy Mechanism}. In \bibinfo{booktitle}{\emph{Proceedings of the
  56th Annual Meeting of the Association for Computational Linguistics (Volume
  1: Long Papers)}}, Vol.~\bibinfo{volume}{1}. \bibinfo{pages}{506--514}.
\newblock


\bibitem[\protect\citeauthoryear{Zhang, Zhang, and Fu}{Zhang
  et~al\mbox{.}}{2017}]%
        {zhang-etal-2017-end}
\bibfield{author}{\bibinfo{person}{Meishan Zhang}, \bibinfo{person}{Yue Zhang},
  {and} \bibinfo{person}{Guohong Fu}.} \bibinfo{year}{2017}\natexlab{}.
\newblock \showarticletitle{End-to-End Neural Relation Extraction with Global
  Optimization}. In \bibinfo{booktitle}{\emph{Proceedings of the 2017
  Conference on Empirical Methods in Natural Language Processing}}.
  \bibinfo{publisher}{Association for Computational Linguistics},
  \bibinfo{address}{Copenhagen, Denmark}, \bibinfo{pages}{1730--1740}.
\newblock


\bibitem[\protect\citeauthoryear{{Zheng}, {Wang}, {Bao}, {Hao}, {Zhou}, and
  {Xu}}{{Zheng} et~al\mbox{.}}{2017}]%
        {zheng2017joint}
\bibfield{author}{\bibinfo{person}{Suncong {Zheng}}, \bibinfo{person}{Feng
  {Wang}}, \bibinfo{person}{Hongyun {Bao}}, \bibinfo{person}{Yuexing {Hao}},
  \bibinfo{person}{Peng {Zhou}}, {and} \bibinfo{person}{Bo {Xu}}.}
  \bibinfo{year}{2017}\natexlab{}.
\newblock \showarticletitle{Joint Extraction of Entities and Relations Based on
  a Novel Tagging Scheme}. In \bibinfo{booktitle}{\emph{Proceedings of the 55th
  Annual Meeting of the Association for Computational Linguistics (Volume 1:
  Long Papers)}}, Vol.~\bibinfo{volume}{1}. \bibinfo{pages}{1227--1236}.
\newblock


\bibitem[\protect\citeauthoryear{{Zhou}, {Su}, {Zhang}, and {Zhang}}{{Zhou}
  et~al\mbox{.}}{2005}]%
        {zhou2005exploring}
\bibfield{author}{\bibinfo{person}{GuoDong {Zhou}}, \bibinfo{person}{Jian
  {Su}}, \bibinfo{person}{Jie {Zhang}}, {and} \bibinfo{person}{Min {Zhang}}.}
  \bibinfo{year}{2005}\natexlab{}.
\newblock \showarticletitle{Exploring Various Knowledge in Relation
  Extraction}. In \bibinfo{booktitle}{\emph{Proceedings of the 43rd Annual
  Meeting of the Association for Computational Linguistics (ACL'05)}}.
  \bibinfo{pages}{427--434}.
\newblock


\bibitem[\protect\citeauthoryear{{Zou}, {Yu}, {Kumar}, and {Wang}}{{Zou}
  et~al\mbox{.}}{2018}]%
        {zou2018unsupervised}
\bibfield{author}{\bibinfo{person}{Yang {Zou}}, \bibinfo{person}{Zhiding {Yu}},
  \bibinfo{person}{B.~V. K.~Vijaya {Kumar}}, {and} \bibinfo{person}{Jinsong
  {Wang}}.} \bibinfo{year}{2018}\natexlab{}.
\newblock \showarticletitle{Unsupervised Domain Adaptation for Semantic
  Segmentation via Class-Balanced Self-training}. In
  \bibinfo{booktitle}{\emph{Proceedings of the European Conference on Computer
  Vision (ECCV)}}. \bibinfo{pages}{297--313}.
\newblock


\end{thebibliography}

\end{document}